\documentclass[a4paper,fleqn]{cas-sc}

\usepackage[authoryear]{natbib}

\usepackage{float}
\usepackage{ragged2e}
\usepackage{caption}
\usepackage{subcaption}
\usepackage{graphics}
\usepackage{CJKutf8}
\usepackage{booktabs}

\def\tsc#1{\csdef{#1}{\textsc{\lowercase{#1}}\xspace}}
\tsc{WGM}
\tsc{QE}
\tsc{EP}
\tsc{PMS}
\tsc{BEC}
\tsc{DE}

\begin{document}
\let\WriteBookmarks\relax
\def\floatpagepagefraction{1}
\def\textpagefraction{.001}

\shorttitle{Chatbots for Mental Health Support}

\shortauthors{Sabour et~al.}

\title [mode = title]{Chatbots for Mental Health Support: Exploring the Impact of Emohaa on Reducing Mental Distress in China}    
\tnotemark[1]
\tnotetext[1]{This work was supported by the National Science Foundation for Distinguished Young Scholars (with No. 62125604) and the NSFC projects (Key project with No. 61936010 and regular project with No. 61876096). This work was also supported by the Guoqiang Institute of Tsinghua University, with Grant No. 2019GQG1 and 2020GQG0005.}

\author[1]{Sahand Sabour}[type=editor, orcid=0000-0003-1927-3562]
\ead{sahandfer@gmail.com}
\author[2]{Wen Zhang}[type=editor, orcid=0000-0002-9405-3167]
\author[3]{Xiyao Xiao}[type=editor, orcid=0000-0002-6326-2206]
\author[3]{Yuwei Zhang}[type=editor]
\author[3]{Yinhe Zheng}[type=editor]
\author[1]{Jiaxin Wen}[type=editor]
\author[4]{Jialu Zhao}[type=editor]
\cormark[1]
\ead{jialuzhao@tsinghua.edu.cn}
\author[1,3]{Minlie Huang}[type=editor]
\cormark[1]
\ead{aihuang@tsinghua.edu.cn}

\cortext[cor1]{Corresponding author}

\affiliation[1]{
    organization={The CoAI Group, DCST, Institute for Artificial Intelligence, State Key Lab of Intelligent Technology and Systems, Beijing National Research Center for Information Science and Technology},
    addressline={Tsinghua University}, 
    city={Beijing},
    citysep={},
    postcode={100084}, 
    country={China}}
    
\affiliation[2]{
    organization={Department of Psychology},
    addressline={Beijing Normal University}, 
    city={Beijing},
    citysep={},
    postcode={100875}, 
    country={China}}
    
\affiliation[3]{
    organization={Beijing Lingxin Intelligent Technology CO., Ltd,},
    addressline={Block D, Yousheng Building, Haidian District}, 
    city={Beijing},
    citysep={},
    postcode={100083}, 
    country={China}}
    
\affiliation[4]{
    organization={Center for Counseling and Psychological Development Guidance Center},
    addressline={Tsinghua University}, 
    city={Beijing},
    citysep={},
    postcode={100084}, 
    country={China}}

\begin{highlights}
\item This study analyzes the effectiveness, acceptability, and practicality of Emohaa, a Chinese conversational agent for mental health support, in reducing mental health distress. 
\item Emohaa provides template-based intervention based on Cognitive Behavioral Therapy (CBT) principles. 
It also provides emotional support by allowing users to discuss their desired topic freely and vent about their problems.
\item Experimental results demonstrate that using Emohaa significantly reduces symptoms of mental distress, namely depression, anxiety, insomnia, and negative affect.
\item Our findings suggest that allowing participants to have open conversations with the agent and receiving emotional support has a complementary effect on the improvements.
\item Based on the results, we conclude that Emohaa is a feasible and effective tool for reducing mental distress. 
\end{highlights}

\begin{abstract}
The growing demand for mental health support has highlighted the importance of conversational agents as human supporters worldwide and in China. 
These agents could increase availability and reduce the relative costs of mental health support. 
The provided support can be divided into two main types: cognitive and emotional support.
Existing work on this topic mainly focuses on constructing agents that adopt Cognitive Behavioral Therapy (CBT) principles. Such agents operate based on pre-defined templates and exercises to provide cognitive support.
However, research on emotional support using such agents is limited.
In addition, most of the constructed agents operate in English, highlighting the importance of conducting such studies in China.
In this study, we analyze the effectiveness of Emohaa in reducing symptoms of mental distress.
Emohaa is a conversational agent that provides cognitive support through CBT-based exercises and guided conversations. It also emotionally supports users by enabling them to vent their desired emotional problems.
The study included 134 participants, split into three groups: Emohaa (CBT-based), Emohaa (Full), and control.
Experimental results demonstrated that compared to the control group, participants who used Emohaa experienced considerably more significant improvements in symptoms of mental distress.
We also found that adding the emotional support agent had a complementary effect on such improvements, mainly depression and insomnia.
Based on the obtained results and participants' satisfaction with the platform, we concluded that Emohaa is a practical and effective tool for reducing mental distress.
\end{abstract}


\maketitle

\section{Introduction}
Mental health is a prevalent issue in the modern world due to the increasing morbidity of mental diseases \citep{who2022}. 
During the COVID-19 pandemic, depression, anxiety, and other mental health issues have increased significantly \citep{2020Prevalence}. 
Specifically, a review by \citet{2020Prevalence} highlighted a 20\% and 35\% rise in depression and anxiety, respectively, for 113,285 individuals across 16 studies. 
Additionally, an international study with a sample of 22,330 adults showed that about 17.4\% of the participants met the criteria for a probable insomnia disorder \citep{2021insomnia}. 
These mental health issues impact people's daily lives, leading to social dysfunction and risks of self-harm and suicide \citep{2020Anxiety}. 
Due to the rapidly increasing demands, mental health services worldwide face challenges regarding lack of professional training and stigmatization of mental illness. 
These challenges can lead to low diagnosis accuracy and patient treatment delays \citep{2020Prevalence}.  

Similarly, the prevalence of mental health diseases in China is increasing. According to the epidemiological survey of mental disorders in China, the lifetime prevalence rate of mental disorders in adults, excluding senile dementia, is 16.57\%. 
Specifically, the prevalence of anxiety disorder is the highest in China, with a 12-month prevalence rate of 4.98\% \citep{2019Prevalence}. 
However, only a limited number of these patients are receiving appropriate treatment for two main reasons.
First, patients lack the mental health knowledge and awareness to take the initiative to seek medical advice.
The second reason is the poor accessibility of mental health services in China. 
As \citet{2019Prevalence}'s epidemiological survey noted, the quality and quantity of mental health services in China are inadequate.

Advancements in Artificial Intelligence (AI) and the field of Natural Language Processing (NLP) have highlighted the potential of machines to serve as anthropomorphic conversational agents.
One of the essential applications of such agents is health care, mainly for providing mental health support.
An early example in this domain is ELIZA \citep{eliza}, which was made to mimic a psychologist and generated responses based on a set of hand-crafted rules.
Employing machines for such tasks increases availability while reducing the costs of seeking support, as these agents could be widely accessible and affordable through mobile devices \citep{fitzpatrick_darcy_vierhile_2017}.
Previous work has shown that individuals are willing to self-disclose their emotional problems with machines \citep{gratch2013using, Ho2018, Liang2021}, 
which is significant as users' self-disclosure is essential for providing support.
It demonstrates user rapport with these agents and highlights their potential as practical and beneficial supporters, thus serving as a strong motivation for this study.

To address their issues with mental health, individuals seek support from others to obtain comfort, reassurance, and a new perspective on their situation \citep{avatar-2022-pauw}.
As proposed by \citet{rime2009}, there are two main types of support to reduce mental distress: cognitive and emotional.
Cognitive support enables individuals to reassess their situation from a different perspective and realize a new way of thinking about their problem \citep{rime2012, avatar-2022-pauw}. 
In contrast, emotional support includes providing validation and understanding to cause relief and improve emotional distress \citep{hill2009helping, rime2012, liu2021esc}.
While cognitive and emotional support is essential for reducing mental distress, as mentioned, the number of human supporters is limited, and many cannot afford the cost of professional help.  
Hence, the invention of high-technology tools or treatments is essential as it can provide effective, available, and affordable support for improving individuals' mental health.

Recent work has mainly focused on delivering cognitive support through conversational agents adopting Cognitive Behavioral Therapy (CBT) principles and has demonstrated the efficacy of such interventions in reducing users' mental distress, mainly depression and anxiety \citep{fitzpatrick_darcy_vierhile_2017, vitalk, inkster_sarda_subramanian_2018, chatbot-2022-liu}. 
In contrast, research on machine-based emotional support is considerably limited.
\citet{liu2021esc} constructed a dataset of emotional support conversations based on \citet{hill2009helping}'s helping skills and demonstrated the feasibility of machine-based emotional support.
Although their work facilitated the research in this direction, it did not conduct an empirical study of the effectiveness of a prototype agent in reducing users' mental distress.
\citet{avatar-2022-pauw} presented the first study investigating the effects of different types of machine-based support, including emotional support.
However, their proposed prototypes only produced a set of pre-defined statements (e.g., "I am sorry to hear that") rather than generating responses based on the users' messages.
Moreover, extensive research on conversational agents in this domain is conducted in English \citep{valizadeh_parde_2022}.
Therefore, with mental health being a rising issue in the Chinese community, existing high demands for available and affordable support in China, and the limited research in this area, we believe constructing and conducting a study on a conversational agent for support in Chinese is crucial.


This study investigates the efficacy of conversational agents for providing cognitive and emotional support.
Specifically, it aims to study the effectiveness of agents providing different types of support in reducing mental distress and assess the acceptability and practicality of such interventions for mental health support.
We introduce Emohaa, a hybrid system involving a platform based on CBT principles and exercises for cognitive support and a conversational platform for emotional support regarding various topics.
We recruit participants from mainland China and hypothesize that frequent use of Emohaa, which includes completing daily exercises and emotional venting, would improve their symptoms of mental distress.


\section{Methods}
\subsection{Emohaa}
Our proposed conversational agent consists of two platforms.
First, a template-based platform that contains conversations with pre-defined options and exercises that assist participants in improving their mental distress based on CBT principles (CBT-based). 
Second, a generative dialogue platform that allows conversations regarding various emotional issues in an open-ended manner (i.e., without requiring the users to choose predefined conversational options) and provides emotional support (ES-based).
\subsubsection{CBT-based}
Creating a platform based on CBT principles postulates a direct and reciprocal interaction between thoughts, feelings, and behaviors that helps illuminate understanding of one's overall emotional distress and situational responses while highlighting areas for intervention \citep{2011Cognitive}. 
As a tool for cognitive support, we integrated two different practices: automatic thoughts training and guided expressive writing.  
Individuals have automatic thoughts in response to a trigger, often outside of that one's conscious awareness. 
These thoughts could often be irrational and harmful when associated with mental distress \citep{1980Cognitive}. 
As one of the core elements of CBT \citep{1998Cognitive,2011Cognitive}, automatic thoughts training aims to identify and dismantle these thoughts (i.e., replace negative thoughts with rational perspectives), which could reduce mental distress and improve one's mood \citep{2011Cognitive, 2000THE}.
In addition, previous studies have shown that writing about stressful or emotional events improves physical and psychological health in non-clinical and clinical populations \citep{Baikie2005Emotional,2007Expressive}.
Therefore, we adopted over 20 guided expressive writing exercises that cover a variety of topics and instruct users throughout each step of the exercise via interactive conversations. 
More details on the exercise design and screenshots of the user interface for this platform are provided in the appendix (Section \ref{sec:appendix} and Figures \ref{fig:cbt-first} \& \ref{fig:cbt-at}).

\subsubsection{ES-based}
Several studies have shown that emotional support is beneficial for reducing mental and emotional distress \citep{burleson2003emotional, heaney2008social, zheng2022augesc}.
In addition, allowing users to discuss their desired topics freely is crucial for creating anthropomorphic conversational agents. 
Therefore, we aimed to construct an agent that could openly converse with users about their emotional problems and generate responses based on their situation.
To this end, we found \cite{liu2021esc}'s dataset of emotional support conversations (ESConv) suitable for this study.
This dataset was constructed based on the Helping Skills Theory \citep{hill2009helping}, in which trained human supporters leverage appropriate support strategies (e.g., self-disclosure, affirmation, and suggestions) to provide adequate emotional support. As the original dataset was curated in English, the conversations were translated into Chinese for the purpose of this study.

For building the ES-based platform, a large-scale Chinese dialogue model was leveraged as the backbone to build a strategy-controlled emotional support dialogue model. 
Specifically, the model chooses an appropriate support strategy given the conversation history.
Accordingly, it generates responses that are coherent with the user's messages and conform to the chosen strategy.
Given the free-flow design of this platform, as opposed to users choosing pre-defined options for the conversation, an additional model \citep{roberta} was trained to classify whether users' messages demonstrated signs of suicidal thoughts to ensure users' safety. 
As individuals with the risk of suicide require immediate professional help, the platform recommends contact information of relevant authorities when corresponding signs are detected. Example conversations with this platform are demonstrated in Figure \ref{fig:es-chat}.

\subsection{Measures}\label{sec:measures}
\subsubsection{PHQ-9}
Participants' depression was measured with the Patient Health Questionnaire (PHQ-9) \citep{kroenke_spitzer_2002}, the most widely used measure in psychological depression trials \citep{fitzpatrick_darcy_vierhile_2017, glischinski_2021}. 
PHQ-9 is a 9-item self-report questionnaire that measures the frequency and severity of depressive symptoms over the last two weeks. 
Participants were asked to score each item from 0 (not at all) to 3 (nearly every day).
By summing the scores from all items, a diagnosis of participants' depression could be made based on the following ranges: 0-5 (no symptoms); 5-9 (mild depression); 10-14 (moderate depression); 15-20 (moderately severe depression); and 20 (severe depression). 
In this study, the internal reliability of the scale (Cronbach's alpha) in the pre-test and the post-test were .78 and .85.

\subsubsection{GAD-7}
To measure participants’ anxiety, we adopted the Generalized Anxiety Disorder (GAD-7) scale \citep{spitzer_2006}, a 7-item questionnaire assessing the frequency and severity of symptoms, thoughts, and related behaviors to anxiety within the last two weeks. 
Like PHQ-9, participants were required to score each item from 0 (not at all) to 3 (nearly every day), with 5, 10, and 15 indicating the cut points for mild, moderate, and severe anxiety ranges, respectively.
The Cronbach’s alpha of this scale in the pre-test and the post-test were both .84. 

\subsubsection{PANAS}
Participants' affect was measured by \citet{watson_clark_tellegen_1988}'s 20-item Positive and Negative Affect Schedule (PANAS). 
In this questionnaire, half of the items represent positive affect (e.g., active, enthusiastic, and proud), and the remaining half corresponds to negative affect (e.g., upset, guilty, and irritable). 
All items are scored on a 5-Likert scale, and higher scores indicate higher levels of affect. 
The Cronbach's alpha of the positive affect dimension in the pre-test and the post-test were .88 and .82, and .85 and .82 for the negative affect dimension at the two-time points. 

\subsubsection{ISI}
The 7-item Insomnia Severity Index (ISI; \cite{morin_1993}) was used to measure participants’ perception of their insomnia. 
This questionnaire assesses the severity of sleep-onset and maintenance difficulties, their interference with daily functioning, and the degree of distress caused by sleep problems. Participants were asked to score each item from 1 (none) to 4 (very severe).
The diagnosis could indicate no insomnia (0-7), sub-threshold insomnia (8–14), moderate insomnia (15–21), and severe insomnia (22–28) based on the sum of scores on all items.
The Cronbach’s alpha of the scale in the pre-test and the post-test were both .87.

\subsection{Participants}
The following criteria were used to recruit participants through online posters: participants were required to be at least 18 years old, able to use a smartphone, not currently in therapy as it would interfere with our study, and not suffering from physical issues as they might influence their psychological state. 
A total of 412 participants registered for the intervention, and 301 met all the above criteria. 
We randomly assigned these participants to three groups: Emohaa (CBT-based), Emohaa (Full), and the control group. 

As mentioned, the ES-based platform allows users to send their desired text messages and employs a generative model for producing its responses, as opposed to the template-based conversations (i.e., providing users with limited conversational options and producing pre-defined answers) in the CBT-based platform.
Due to the existing limitations of generative models, such as problems with response coherence and fluency, we believed a direct comparison between the effectiveness of the two platforms was inappropriate.
Therefore, we required participants in the Emohaa (Full) group to use both platforms and aimed to study the complementary effect of the ES-based platform rather than analyzing its respective efficacy.
In addition, considering the relatively long waiting time and the potential number loss in the control group, we allocated 30 more participants to the control group.
Accordingly, a research assistant informed the participants of our code of conduct and asked for consent through WeChat, China's most popular social media platform. 
Participants were assured that their participation was voluntary and anonymous.

\begin{figure*}[!ht]
\centering
\includegraphics[width=.9\linewidth]{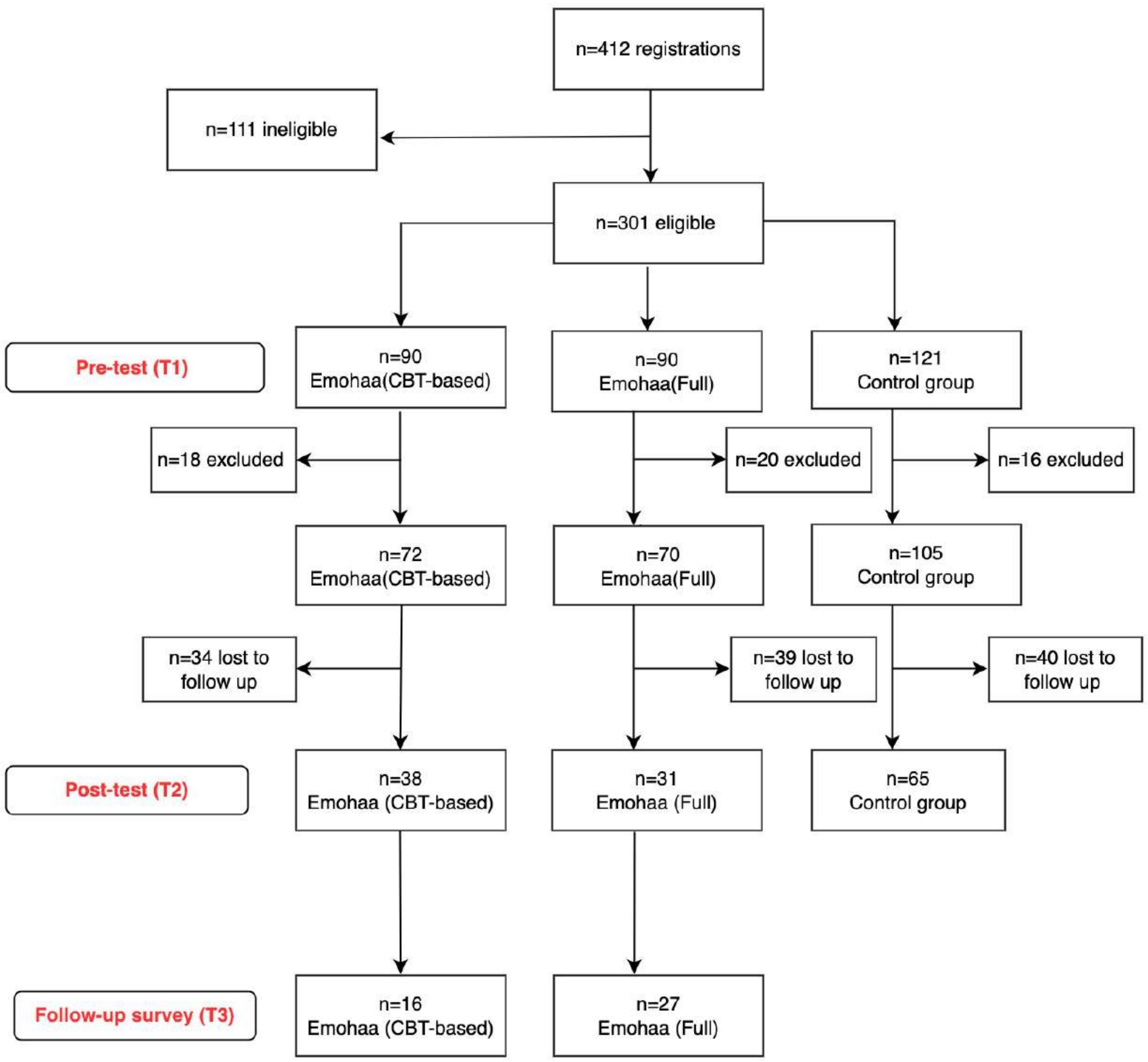}
\caption{Flowchart of the participant recruitment process.}
\label{fig:recruit}
\end{figure*}

After signing the consent form, participants were instructed to take the pre-test (T1) questionnaires, including PHQ-9, GAD-7, PANAS, and ISI (section \ref{sec:measures}), and their demographic information. 
Sixteen participants were excluded from the study and referred to relative authorities for professional help as they were at risk of suicide according to the PHQ-9, and 38 participants were excluded because of more than half of the missing data.
Overall, 72 participants in the Emohaa (CBT-based), 70 participants in the Emohaa (Full), and 105 participants in the control group completed the pre-test questionnaires. 
Similarly, one day after the end of the intervention, all the participants were asked to fill in the post-test (T2) questionnaire, which included the same items as T1. 
Additionally, one month after the end of the experiment, participants were invited to fill in a follow-up questionnaire (T3) intervention to track the lasting effect of the intervention. 
From the perspective of health ethics and practical reasons, other forms of intervention could have been provided to the control group after the intervention.
Hence, there were no valid data at T3 for the control group. 
The above recruitment process is illustrated in Figure \ref{fig:recruit}.

Of the randomized participants, 54.2\% (134/247) went on to provide partial or complete data at T2. 
Independent t-tests analyses did not detect evidence of significant differences at T1 between those who dropped out of the study versus those who did not on age (\textit{t} $=1.51$; \textit{p} $=.132$); gender (\textit{t} $=.37$; \textit{p} $=.709$); working tenure (\textit{t} $=.92$; \textit{p} $=.357$); PHQ-9 (\textit{t} $=.95$; \textit{p} $=.342$); GAD-7 (\textit{t} $=.59$; \textit{p} $=.558$); PANAS of positive (\textit{t} $=1.02$; \textit{p} $=.145$) and negative (\textit{t} $=1.21$; \textit{p} $=.227$) affect scores; or on insomnia (\textit{t} $=1.17$; \textit{p} $=.244$).

\subsection{Data Collection and Privacy}

\subsubsection{Guidelines}
All participants were required to complete the mental distress questionnaires (Section \ref{sec:measures}) at T1 and T2.
Excluding the control group, all participants were instructed to use the CBT-based platform daily, which required completing at least one automatic thinking exercise and writing a guided expressive diary.
However, participants were encouraged to complete more exercises for better outcomes.
In addition, participants from the Emohaa (Full) group were tasked to converse with the ES-based platform at least once daily.
Each conversation session was required to last for 5-10 conversational turns. Participants were encouraged to continue chatting with the platform if they felt engaged in the conversation.
Although there were no limitations on the conversational topics, participants were encouraged to talk about two main types of emotional experiences and problems: Event-based (e.g., breaking up with a partner, problems with work/study, and nuisance complaints); Emotion-based (i.e., topics that cause anger, sadness, anxiety).

\subsubsection{Quality Control}
Participants' usage of the platform was manually checked every three days. 
Those who failed to conform to the guidelines were notified and required to complete the relative tasks to ensure high adherence.
In addition, conversations with the ES-based platform were analyzed to monitor the chatbot's performance and the reliability of the conversations. 
For instance, during the first check of this intervention, it was found that 3/34 participants sent the same message multiple times to meet the requirements.
These participants were contacted and asked to repeat these conversations to ensure the experiment's integrity.

\subsection{Privacy and Ethics Statement}
Regarding the conversations with Emohaa, participants were instructed not to share any personal information (e.g., name, address, and date of birth) that could be used to identify them.
The data collected during the experiment are anonymized, stored securely, and will be available for research purposes through a request to the corresponding author. 
This study was approved by our Institutional Review Board (IRB). 
Participants were informed of the purpose of this study and consented to be part of this experiment.

\section{Results}
\subsection{User Demographics}
Demographic information of our studied sample ($n=134$) is provided in Table \ref{table:user_demo}. 
Overall, the majority of participants were female (107/134, 79.85\%). 
The average age of the studied sample was 31.81 years old (\textit{SD} $= 8.96$).
Participants had worked for an average of 9.10 years (\textit{SD} $= 9.86$) prior to the experiment.
All of the participants were from Mainland China.
As the baseline for participants' mental distress, on average, the samples showed mild ranges of depression (\textit{Mean} $=8.01$, \textit{SD} $= 4.18$), mild anxiety (\textit{Mean} $=8.16$, \textit{SD} $= 3.94$), and sub-threshold insomnia (\textit{Mean} $=9.78$, \textit{SD} $= 4.4$).
In regards to PANAS, participants on average demonstrated moderate levels of positive (\textit{Mean} $=24.75$, \textit{SD} $= 6.5$) and negative affect (\textit{Mean} $=21.97$, \textit{SD} $= 6.78$).

\begin{table*}[!ht]
\caption{
    User Demographics of Our Studied Sample ($n=134$).
  }
  \label{table:user_demo}
    \begin{tabular}{c c c c c c}
        \toprule
        \multicolumn{2}{c}{} &
        \multicolumn{1}{c}{Emohaa} & 
        \multicolumn{1}{c}{Emohaa} & 
        \multicolumn{1}{c}{Control} &
        \\
        \multicolumn{2}{c}{} &
        \multicolumn{1}{c}{CBT-based} & 
        \multicolumn{1}{c}{Full} & 
        \multicolumn{1}{c}{Group} &
        \\
        \multicolumn{2}{c}{} &
        \multicolumn{1}{c}{($n=38$)}& 
        \multicolumn{1}{c}{($n=31$)} & 
        \multicolumn{1}{c}{($n=65$)} &
        \\ \midrule
        Gender, $n$ (\%) & Male & 6 (15.79\%) & 9 (29.03\%) & 12 (18.46\%)  
        \\ 
         & Female &32 (84.21\%)& 22 (70.97\%)& 53 (81.54\%)  \\
         Age, \textit{Mean (SD)}  &  &  31.18 (7.94)&28.55 (7.21) & 33.74 (9.84)\\
         Work Experience, \textit{Mean (SD)} &  &8.26 (8.15)& 6.03 (6.89)&11.05 (11.51) \\
        \bottomrule
    \end{tabular}
\end{table*}

\subsection{Effects of Emohaa Intervention}

\begin{table*}[ht]
  \caption{
 Analyses Results of Variance in Mental Health Outcomes.
  }
  \label{table:mental_results}
    \begin{tabular}{c c c c c c c c c c c}
        \toprule
        \multicolumn{2}{c}{} &
        \multicolumn{2}{c}{Emohaa} & 
        \multicolumn{2}{c}{Emohaa} & 
        \multicolumn{2}{c}{Control} &
        \multicolumn{3}{c}{}\\
        \multicolumn{2}{c}{} &
        \multicolumn{2}{c}{CBT-based} & 
        \multicolumn{2}{c}{Full} & 
        \multicolumn{2}{c}{Group} &
        \multicolumn{3}{c}{}\\
        \multicolumn{2}{c}{} &
        \multicolumn{2}{c}{($n=38$)}& 
        \multicolumn{2}{c}{($n=31$)} & 
        \multicolumn{2}{c}{($n=65$)} &
        \multicolumn{3}{c}{}
        \\
        \cmidrule{3-8}
        Variables & Time  &\textit{Mean} & \textit{SD} & \textit{Mean}  & \textit{SD} &\textit{Mean}  & \textit{SD} & $F^a$& $p$ & $\eta ^ 2$\\ \midrule
        Depression & 1 & 12.97 & 4.10 & 12.61 & 4.40 &  11.17 & 4.01 & 19.11 & < .001 &  0.23 \\ 
         &2 & 10.13 & 2.96 &  9.23 & 3.43 &  12.49 & 5.21 & - &  - &  - \\ 
        Anxiety & 1 & 16.13 & 3.54 &  15.45 & 3.75 &  14.45 & 4.15 & 12.01 &  < .001 &  0.16 \\ 
          &2 & 12.53 & 3.29 &  12.13 & 3.66 &  14.26 & 4.03 & - &  - &  - \\ 
        Positive affect & 1 & 23.55 & 6.18 &  24.19 & 5.69 &  25.71 & 7.05 & 1.85 &  0.161 &  0.03 \\ 
          &2 & 26.58 & 6.06 &  27.48 & 5.93 &  26.9 & 6.94 & - &  - &  - \\ 
        Negative affect & 1 & 23.45 & 6.22 &  22.35 & 8.13 &  20.92 & 6.3 & 12.11 &  < .001 &  0.16 \\ 
          &2 & 17.71 & 5.18 &  19.1 & 6.36 &  21.09 & 7.36 & - &  - &  - \\ 
        Insomnia & 1 & 17.32 & 5.79 &  18.74 & 5.63 &  16 & 5.83 & 3.52 &  0.033 &  0.05 \\ 
          &2 & 15.18 & 4.99 &  16.52 & 5.51 &  16.12 & 6.37 & - &  - &  - \\ \bottomrule
    \end{tabular}\par

\justify{\textbf{Note} The F-test tests the Condition × Time interaction effect to detect a significant difference among the conditions in the rate of change across time.}
\end{table*}

\begin{figure*}
\centering
\begin{subfigure}{.48\linewidth}
    \centering
    \includegraphics[width=\linewidth]{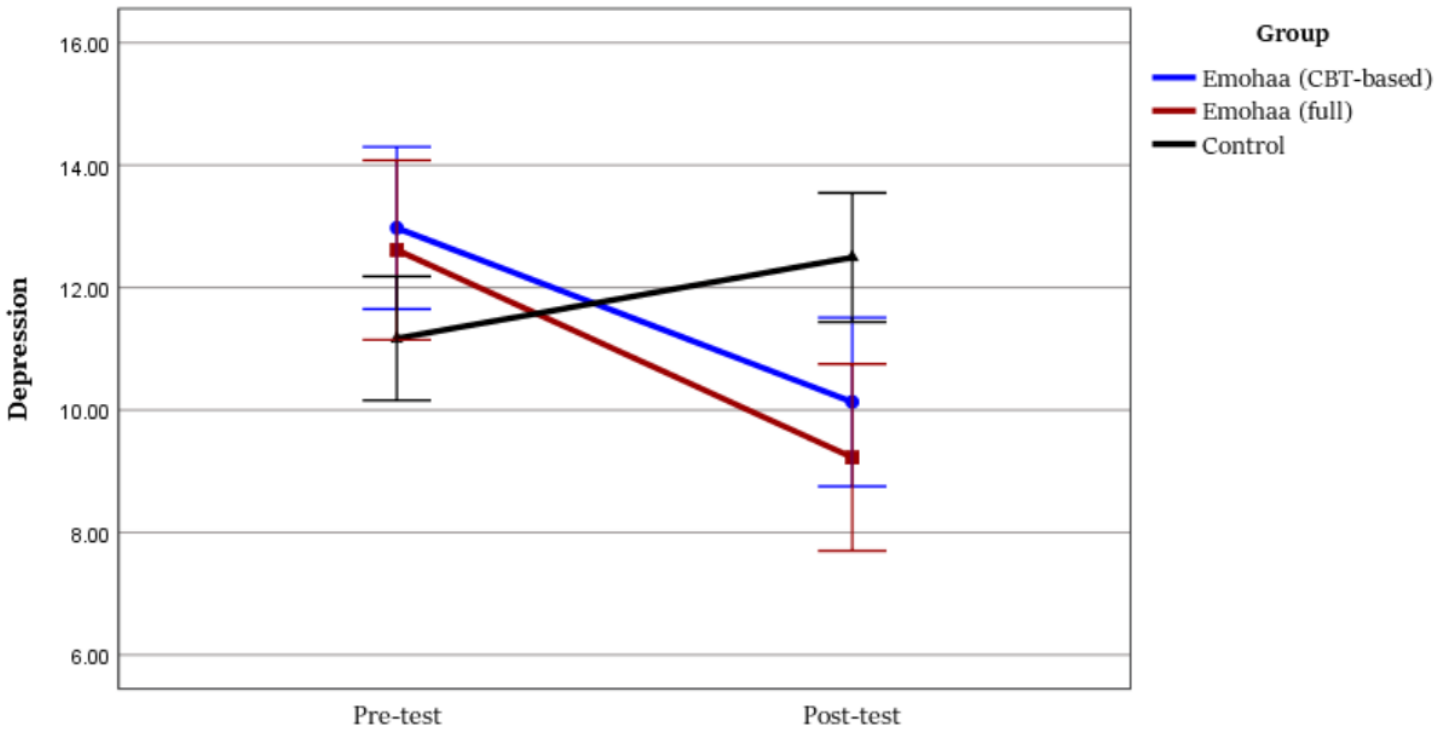}
    \caption{PHQ-9}
    \label{fig:phq9}
\end{subfigure}
    \hfill
\begin{subfigure}{.48\linewidth}
    \centering
    \includegraphics[width=\linewidth]{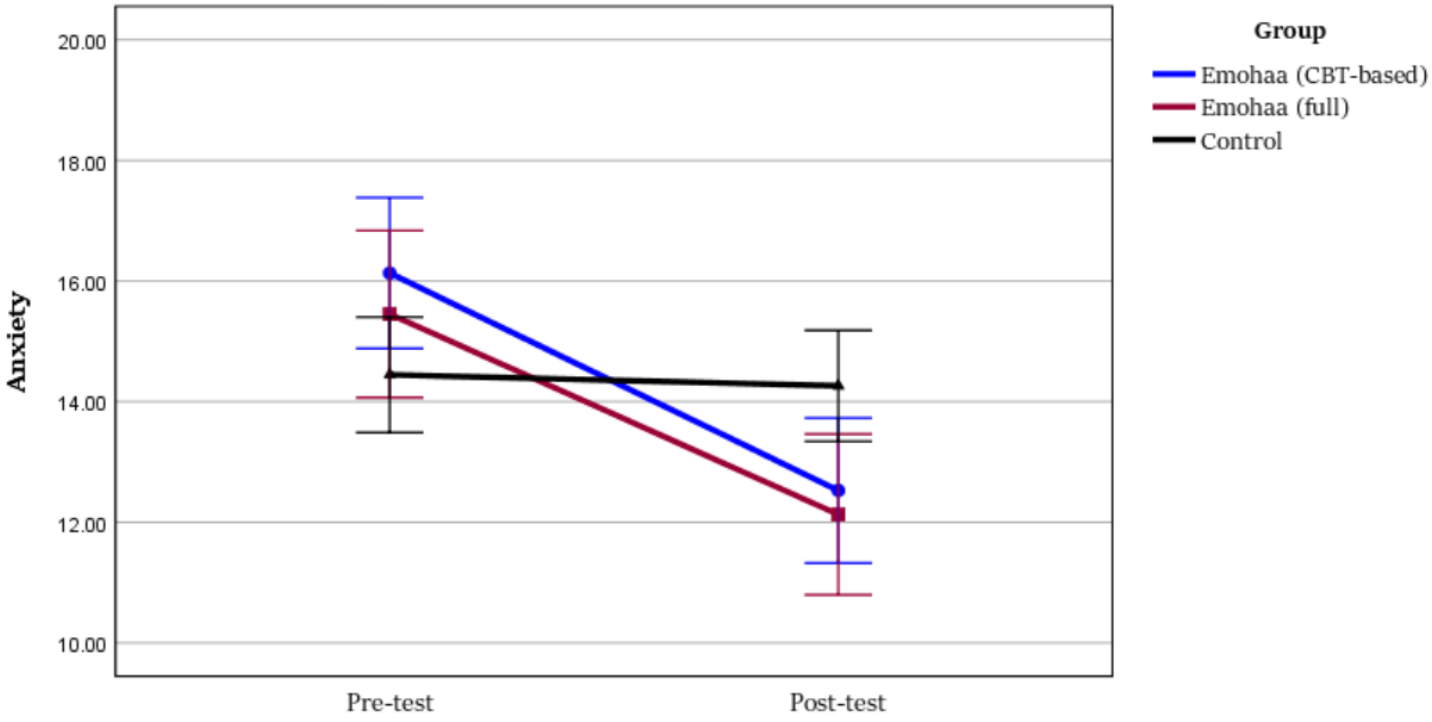}
    \caption{GAD-7}
    \label{fig:gad7}
\end{subfigure}

\bigskip
\begin{subfigure}{.48\linewidth}
    \centering
    \includegraphics[width=\linewidth]{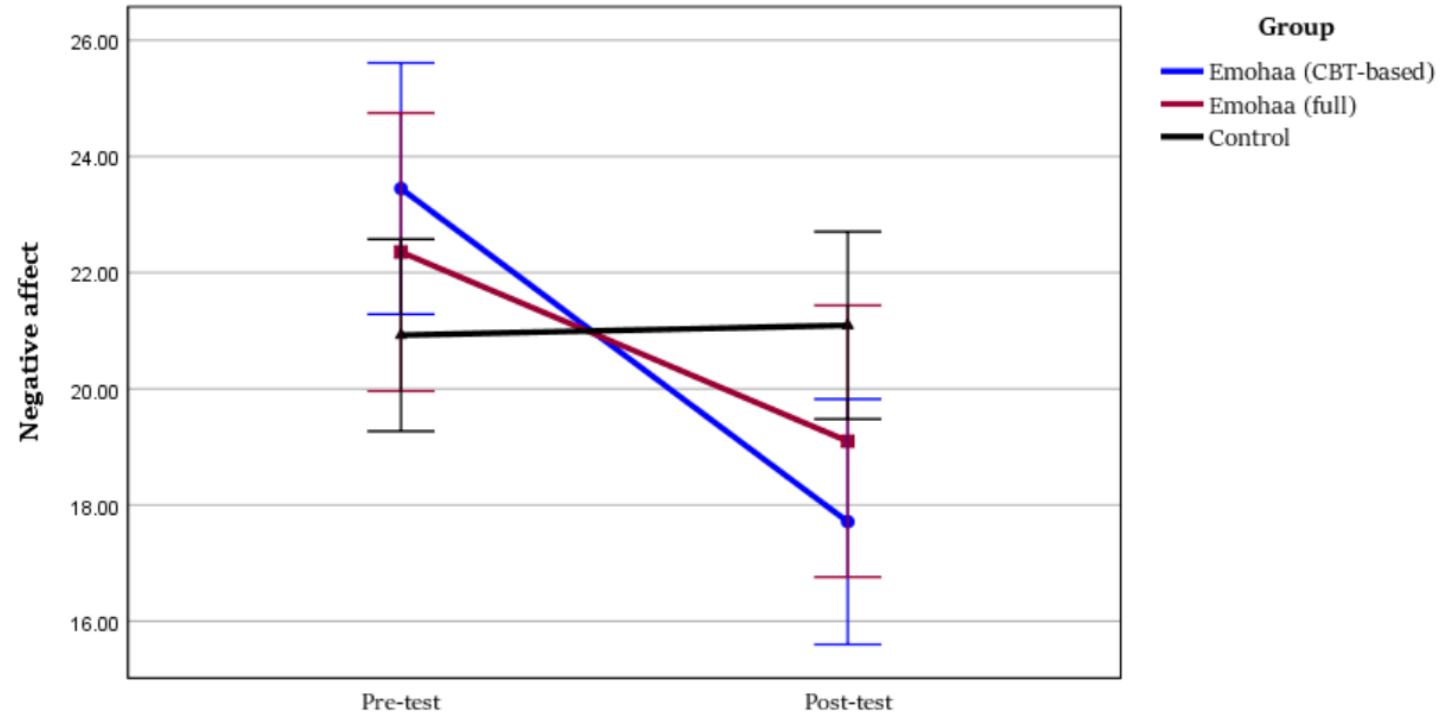}
    \caption{PANAS}
    \label{fig:panas}
\end{subfigure}
    \hfill
\begin{subfigure}{.48\linewidth}
    \centering
    \includegraphics[width=\linewidth]{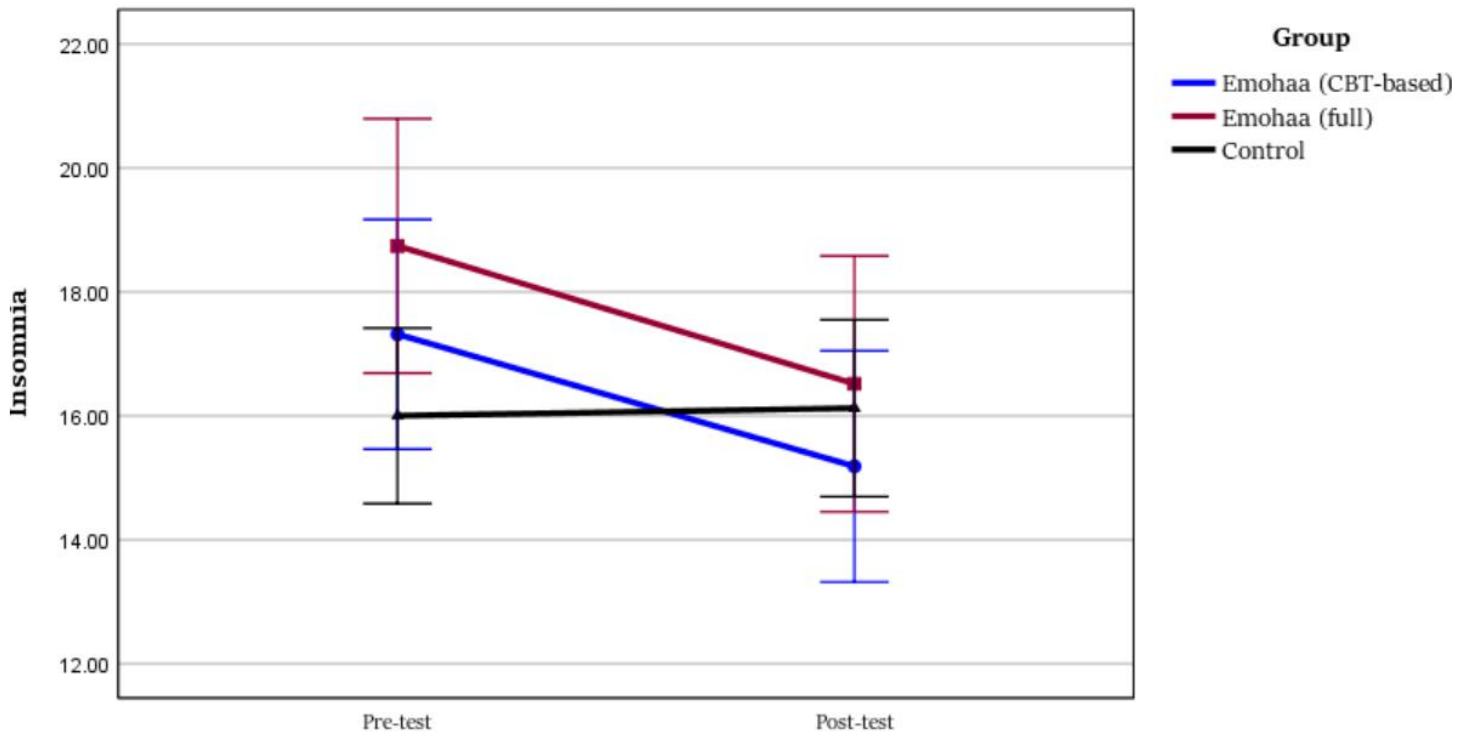}
    \caption{ISI}
    \label{fig:isi}
\end{subfigure}
\caption{Changes in the mean mental distress scores by group over the initial intervention period (T1-T2). Error bars indicate 95\% confidence interval.}
\label{fig:images}
\end{figure*}

To investigate whether the effects of interventions were different from each other, and from that of the control group, we conducted a one-way repeated measures MANOVA with time (two levels: pre-test and post-test) and group type (three levels: Emohaa (CBT-based) vs. Emohaa (Full) vs. Control) as the independent variables and the five mental health indicators as the dependent variables. First, the results showed a main effect of time on depression (\emph{F} [1, 131] = 18.96, \emph{p} < .001, $\eta ^2$ = .13), indicating that participants’ depression decreased over time. Furthermore, as presented in Table 2, there were significant Group × Time interaction effects for depression, \emph{F} [2, 131] = 19.11, \emph{p} < .001, $\eta ^2$ = .23, indicating a significant difference in participants’ depression changes among the three groups, and such difference had a relatively large effect size that is more than .14 \citep{cohen_1988}. Specifically, as Figure \ref{fig:phq9} shows, the intervention effects on depression stemmed from decreases in both Emohaa (CBT-based) (\emph{t} = -4.91, \emph{p} < .001) and Emohaa (Full) group (\emph{t} = -4.05, p < .001) from pre-test to post-test, but there was an increase of depression in the control group (\emph{t} = 2.54, \emph{p} = .013) from pre-test to post-test. Moreover, we did not find significant differences between the two types of interventions (\emph{F} [1, 67] = .30, \emph{p} = .584). 

Similarly, we conducted the same MANOVA analyses to test whether participants’ anxiety changed differently over the intervention period. The results revealed a main effect of time on anxiety, \emph{F} [1, 131] = 45.04, \emph{p} < .001, $\eta ^2$ = 0.26; indicating participants’ anxiety decreased over time. As Table \ref{table:mental_results} shows, the interaction effect of Group × Time was significant on anxiety \emph{F} [2, 131] = 12.00, \emph{p} < .001, $\eta ^2$ = .16. Specifically, as shown in Figure \ref{fig:gad7}, participants’ anxiety decreased significantly in both Emohaa (CBT-based) (\emph{t} = -5.69, \emph{p} < .001) and Emohaa (Full) (\emph{t} = - 4.53, p < .001) groups from pre-test to post-test, but participants’ anxiety did not significantly change in the control group (\emph{t} = -.39, \emph{p} = .397) from pre-test to post-test. No significant differences were found in the effects of the two types of interventions (\emph{F} [1, 67] = .09, \emph{p} = .771). 

Additionally, the results showed that there was a main effect of time on positive affect (\emph{F} [1, 131] = 19.72, \emph{p} < .001, $\eta ^2$ = .13) and on negative affect too (\emph{F} [1, 131] = 29.20, \emph{p} < .001, $\eta ^2$ = .18), meaning that participants’ positive affect increased and negative affect decreased over time. However, we did not find a significant interaction effect of Group × Time on positive affect (\emph{F} [2, 131] = 1.85, \emph{p} = .16, $\eta ^2$ = .03). The interaction effect of Group × Time was significant on negative affect (\emph{F} [2, 131] = 12.11, \emph{p} < .001, $\eta ^2$ = .16.) Specifically, as shown in Figure \ref{fig:panas}, participants’ negative affect decreased significantly in both Emohaa (CBT-based) (\emph{t} = -7.47, \emph{p} < .001) and Emohaa (Full) (\emph{t} = -2.14, \emph{p} = .040) groups from pre-test to post-test, but their negative affect was not significantly changed in the control group (\emph{t} = .26, \emph{p} = .795). The post-hoc results showed that there was no significantly different effects of PA (\emph{F} [1, 67] = .03, \emph{p} = .861) and NA (\emph{F} [1, 67] = 2.36, \emph{p} = .129) between types of interventions.

Finally, the MANOVA results demonstrated a main effect of time on participants’ insomnia (\emph{F} [1, 131] = 9.65, \emph{p} = .002, $\eta ^2$ = .07), indicating that participants’ insomnia improved during the period of intervention. Besides, the results showed a significant interaction effect of Group × Time on insomnia (\emph{F} [2, 131] = 3.52, \emph{p} = .031, $\eta ^2$ = 0.05), but the differences had a relatively small effect size that was between small effect size of .01 and medium effect size of .06 (Cohen, 1988). Specifically, as Figure \ref{fig:isi} reveals, the effects stemmed from a significant insomnia decrease in Emohaa (CBT-based) group (\emph{t} = -3.84, \emph{p} < .001), a marginally significant decrease of insomnia in Emohaa (Full) group (\emph{t} = -2.01, \emph{p} = .053), and no difference in the control group between the pre-test and post-test (\emph{t} = .19, \emph{p} = .869). We did not find a significantly different effects of the two types of interventions (\emph{F} [1, 67] = .02, \emph{p} = .936).

\subsubsection{Supplemental Analyses}\label{sec:supplemental}

We collected participants’ data on the mental health indicators (section \ref{sec:measures}) three weeks after the post-test. Due to practical reasons that participants in the control group received other forms of interventions after the post-test, we only collected two intervention groups’ data. 16 participants in the Emohaa (CBT-based) group and 27 in the Emohaa (Full) group returned the questionnaires. Among the participants, about 25.6\% were male, and 74.4\% were female.  Their average age was 29.40 years old (\emph{SD} = 7.76), and their average working tenure was 6.81 years (\emph{SD} = 7.75). 

To compare the effects between two intervention groups, we conducted MANOVA with time (three levels: pre-test, post-test, and three weeks after post-test) and group type (two levels: CBT-based Emohaa vs. full Emohaa) as the independent variables and the five mental health indicators as the dependent variables. Results showed that there were no significant interaction effects of Time × Group on depression (\emph{F} [1, 41] = 1.26, \emph{p} = .267, $\eta ^2$ = 0.03), anxiety (\emph{F} [1, 41] = .58, \emph{p} = .451, $\eta ^2$ = .01), positive affect (\emph{F} [1, 41] = .41, \emph{p} = .527, $\eta ^2$ = .01) or negative affect (\emph{F} [1, 41] = .44, \emph{p} = .511, $\eta ^2$ = 0.01), indicating that the changes of participants’ four mental health indicators did not vary from each other between the two groups. However, such interaction effect was significant on insomnia (\emph{F} [1, 41] = 7.33, \emph{p} = .010, $\eta ^2$ = .15). The difference stemmed from that participants’ insomnia symptoms returned to the pre-test level in the Emohaa (CBT-based) group. Still, participants’ insomnia in the Emohaa (Full) group continued improving after the intervention. Additionally, we found a main effect of the group on anxiety (\emph{F} [1, 41] = 5.04, \emph{p} = .030, $\eta ^2$ = .11), meaning that although the two groups’ anxiety rose again after the intervention; the Emohaa (Full) group had longer lasting effect on participants’ anxiety than the Emohaa (CBT-based) group. 

\subsection{Conversation Analysis}
\begin{CJK*}{UTF8}{gbsn}
During the experiment, participants had 7 conversation sessions with Emohaa on average (SD = 6.62, Max = 18, Min = 4).
These sessions had a mean of 17 conversational turns (SD = 10.68, Max = 87, Min = 5).
N-gram analysis was used to investigate the characteristics of participants' conversations with Emohaa. 
The most discussed keywords were found to be 感觉 (feeling; 33\%), 工作 (work; 20\%), 心情 (mood; 11\%), 学习 (pressure; 10.5\%), 朋友 (friends; 9.2\%), and 孩子 (children; 7.7\%). 
Percentages indicate the proportion of conversations that included the keyword.
The main problems that participants wanted to talk about were 工作环境 (Work environment), 工作压力 (Work pressure), 浪费时间 (Wasting Time), 集中注意力 (Keeping focus), 牺牲休息时 (Sacrificing leisure time), and 转移注意力 (Diverted attention).
In general, participants were mainly interested in 正念冥想 (Mindful meditation), 早点休息 (Resting early), 身体健康 (Being healthy), and 提供情绪价值 (Providing emotional value).
\end{CJK*}

\subsection{Acceptability and Feasibility}
Participants who had used Emohaa during the experiment were instructed to complete an additional survey to evaluate the agent's performance.
Most participants (60/69, 86.9\%) reported that they had never received psychological counseling before the experiment, and only two had taken psychotropic medication.

Initially, participants were asked to rate the CBT-based platform's ease of use, provided content, and interface quality on a 10-point Likert scale.
Most participants reported moderate to high levels of satisfaction with the platform's functionality (56/69, 81.16\%) and the designed exercises (47/69, 68.12\%).
In addition, more than half of the participants (43/69, 62.32\%) were satisfied with the interface design. 
Overall, the majority (49/69, 71\%) reported that they would recommend this platform to others. 

Similarly, participants who had used Emohaa's ES-based platform were instructed to rate its performance. 
This platform was considered by most of the participants as an appropriate chatting partner (24/31, 77.42\%) and channel for emotional venting (20/31, 64.5\%), and more than half of the participants (18/31, 58.1\%) reported that chatting with this platform made them feel heard.
When asked about their expectations of the platform, the majority believed it to be a suitable companion for emotional companionship and support that can accurately interpret their emotions and provide emotional counseling (21/31, 67.74\%).
In addition, most participants were satisfied with the interface (20/31, 64.5\%) and reported that they would recommend it to others (19/31, 61.3\%). Independent t-tets showed that there was no significant difference between participants' satisfaction with Emohaa CBT-based platform and ES-based platform (\emph{t} = 1.16; \emph{p} = .250).

\begin{table*}[ht]
  \caption{
 Summary of Participants' Feedback on Emohaa.
  }
  \label{table:user_feedback}
    \begin{tabular}{c c c c}
        \toprule
        Platform &
        \multicolumn{2}{c}{Problems} 
        & $n$
        \\ \midrule
        \multirow{7}{*}{CBT-based} & \multirow{2}{*}{Content}  & Repetitiveness & 5\\
        &  & Limited exercises and options& 11\\
        \cmidrule{2-4}
        & \multirow{4}{*}{Design}  & Unclear instructions & 12\\
        &  & Impractical Tips and Recommendations& 6\\
        &  & Technical (i.e., Bugs,  glitches, and lags)& 14\\
         & & Ambiguous exercise tracking and transition  & 5\\
         \midrule
         \multirow{4}{*}{ES-based} & \multirow{2}{*}{Functionality}  & Lack of initiative in conversations& 2\\
         &  & Unable to understand image/video/audio inputs& 3\\
         \cmidrule{2-4}
         & \multirow{2}{*}{Generation}  & Rigid conversations& 10\\
         &  & Unrelated and out-of-context responses& 8\\
         \bottomrule
        
    \end{tabular}
\end{table*}

Lastly, participants were asked to provide feedback on their experience with Emohaa. 
Table \ref{table:user_feedback} summarizes the most common themes in the collected responses.
The most frequently raised concerns were technical issues, unclear instructions, and limited content and choices. 
The reported technical issues were mainly regarding the user interface (e.g., "Cannot click the next page" and "Accidentally closing the app removed all my progress"). 
Many participants were overwhelmed with the number of categories in the guided writing exercises and felt some topics were illogical and not applicable to real life. 
Participants also felt that the stories in the automatic thinking exercises were excessive while there were not enough options to describe their mood and emotions after completing the exercise.
Moreover, it was suggested that the dialogue options in the platform's template were inadequate.

Issues regarding the performance of the ES-based platform were also raised. 
Several participants reported that the conversations were rigid, and the system needed user guidance to continue the conversation. 
In some instances, the generated responses were reported as irrelevant or incoherent to the conversation.
Participants also highlighted the platform's occasional inability to remember what had been said in the early stages of the conversation, initiate conversation topics, and understand various input types (i.e., audio, video, and image).

In addition, participants were also asked to provide suggestions on how to improve Emohaa. 
As concerns regarding lack of content and options were mentioned, it was suggested that additional scenarios, stories, instructions, and options be included in the CBT-based platform. 
The importance of regularly updating the platform and promptly fixing technical issues was also highlighted. 
Regarding the ES-based platform, nearly half of the participants (14/31, 45.1\%) believed that improvements for making the generated responses less rigid were necessary. 
It was also suggested that support for different input types be added to create a more interactive and engaging experience.
Many also believed that recommending mental health-related content during conversations and taking the initiative in conversations would benefit this platform.

\section{Discussion}
\subsection{Main Findings} 
The obtained results demonstrated that users experienced reduced levels of mental distress in the measured categories after using Emohaa. 
Compared to the control group, there was a significant decrease in depression and anxiety of the participants who used Emohaa, as measured by the PHQ-9 and the GAD-7 questionnaires, respectively.
Similarly, as measured by the PANAS and the ISI questionnaires, their negative affect and insomnia were also considerably reduced. 
Therefore, as shown by the experimental results, Emohaa can be seen as an effective tool for mental health support.

Regarding the difference in outcomes between the two groups that used Emohaa, no significant differences were found in the short term.
Both interventions effectively relieved individuals' mental health symptoms. 
However, as provided by the supplemental analyses (section \ref{sec:supplemental}), participants who used the ES-based platform showed comparatively fewer indicators of insomnia.
This finding highlights a potential benefit of emotional venting in improving problems regarding sleep in the long term.
Moreover, although the anxiety indicators of all participants increased at T3, this increase was considerably smaller for those who had access to the ES-based platform.

Based on the obtained feedback, most participants were satisfied with this agent and considered recommending it to others.
In line with previous research \citep{Liang2021, Ho2018}, the results of the conversation analysis indicated that participants were willing to self-disclose their emotional problems, as shown by their most discussed keywords and topics.
Moreover, most participants considered Emohaa's ES-based platform a chatting partner that can effectively listen to their problems and provide a channel for them to vent their emotions.
Notably, the majority felt that this platform could understand their emotions, an essential feature of conversational agents for support and a crucial trait for establishing a therapeutic connection \citep{chatbot-2022-liu}.
Therefore, our findings suggest that Emohaa can also be seen as an acceptable and feasible tool for support.

In addition to highlighting Emohaa's effectiveness in mental health support, this study demonstrated the potential of generative conversational agents and combining emotional and cognitive support to reduce mental distress symptoms.
Our findings suggest that allowing users to freely converse about their desired topics with the agent has a complementary effect when added to more common forms of machine-based support (i.e., template-based conversations and exercises for cognitive support through CBT).

\subsection{Limitations and Future Work}
This study had several limitations regarding its design and methodology.
The study duration was limited; thus, only two assessments of participants' mental distress were made.
Although a follow-up screening for participants that had used Emohaa during the experiment was conducted, no data regarding the control group's participants were gathered as they might have received other interventions after the initial two-week screening.
It is believed that the number of participants was sufficient to demonstrate the preliminary effects of employing conversational agents for mental health support in theory.
However, the sample size and the experiment duration are inadequate for generalizing the obtained results of this study to the public.
Future experiments will include larger sample size and longer study duration to further ensure the generalizability of Emohaa's effectiveness in reducing mental distress.

As mentioned, Emohaa's several technical issues could substantially impact the users' perceived level of empathy and support \citep{fitzpatrick_darcy_vierhile_2017}, so they should be resolved promptly. 
A management system for addressing similar issues on time should also be implemented in future work.
Moreover, several participants raised issues regarding Emohaa's limited content (i.e., exercises and options) and unclear instructions.
Similar to \cite{chatbot-2022-liu}, a wider variety of psychological resources will be consulted in future work to expand the provided content in the CBT-based platform and revise the instructions to avoid user misinterpretations or confusion.

Regarding the ES-based platform, several reported instances suggested that Emohaa forgets the information in previous turns and that the generated responses are irrelevant to the context, which could impair user engagement and rapport. 
This is a well-known issue in current language models \citep{PELAU2021106855}, and the main reason could be the limited number of words in the model's input (128 words for Emohaa).
A feasible approach to address this issue is to add a module that could summarize the essential information of the previous turns in the conversation \citep{blenderbot2}.
In addition, previous work has demonstrated the benefits of adding persona \citep{zheng2019personalized, emily-2021, wu2021transferable} and commonsense knowledge \citep{li2020empathetic, Sabour2022CEM} for improving user experience with generative conversational agents.
Future work could explore these additions to study their efficacy and corresponding improvements in mental health support.

\section{Conclusions}
The present study introduced Emohaa, a Chinese conversational agent for mental health support.
Emohaa employs CBT principles to provide cognitive support through template-based guided conversations for expressive writing and automatic thinking exercises.
In addition, it includes a platform for providing emotional support in which users can discuss their desired emotional problems.
This study examined the effectiveness of Emohaa in reducing mental distress and investigated its feasibility and acceptability as a tool for mental health support in China.
Our findings demonstrated that participants experienced fewer symptoms of mental distress after using Emohaa for the duration of the study.
Hence, we believe this agent could serve as a valuable tool for reducing users' mental distress.
In addition, we found that implementing the generative dialogue platform for emotional support had a complementary effect on improving their mental distress, mainly depression and insomnia. 
This finding highlights the potential of generative conversational agents for the future of mental health support.
In the future, we hope our work can inspire other studies to expand upon our research, leverage generative models for providing support, and investigate their comparative efficacy.

\appendix
\section{User Interface} \label{sec:appendix}
\begin{figure*}[!ht]
\centering
\hspace{1cm}
\begin{subfigure}{.43\linewidth}
    \centering
    \includegraphics[width=\linewidth, height=9cm]{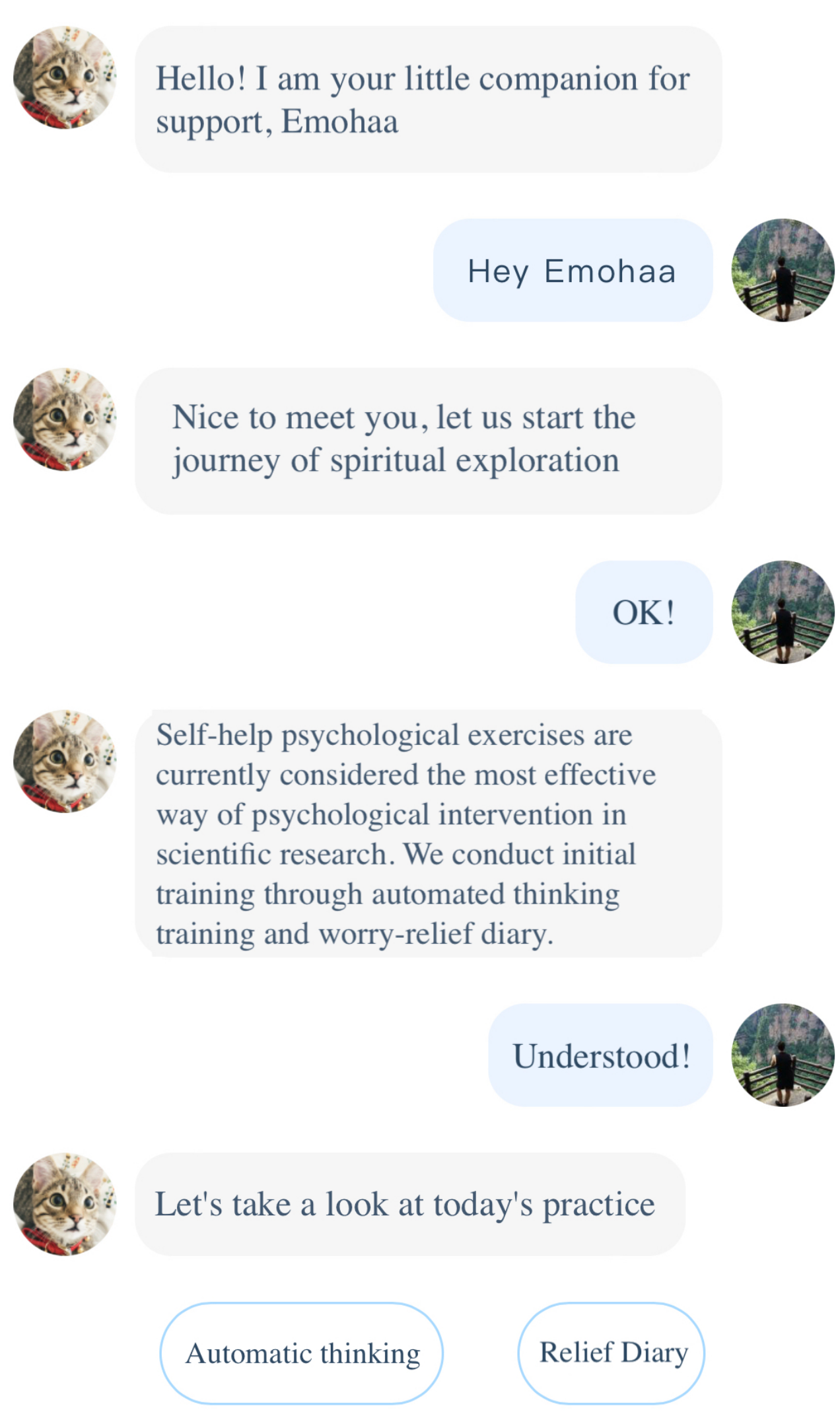}
    \caption{Template-based Conversations for Exercise Guidance}
    \label{fig:cbt_chat}
\end{subfigure}
\hspace{.5cm}
\begin{subfigure}{.4\linewidth}
    \centering
    \includegraphics[width=\linewidth, height=9cm]{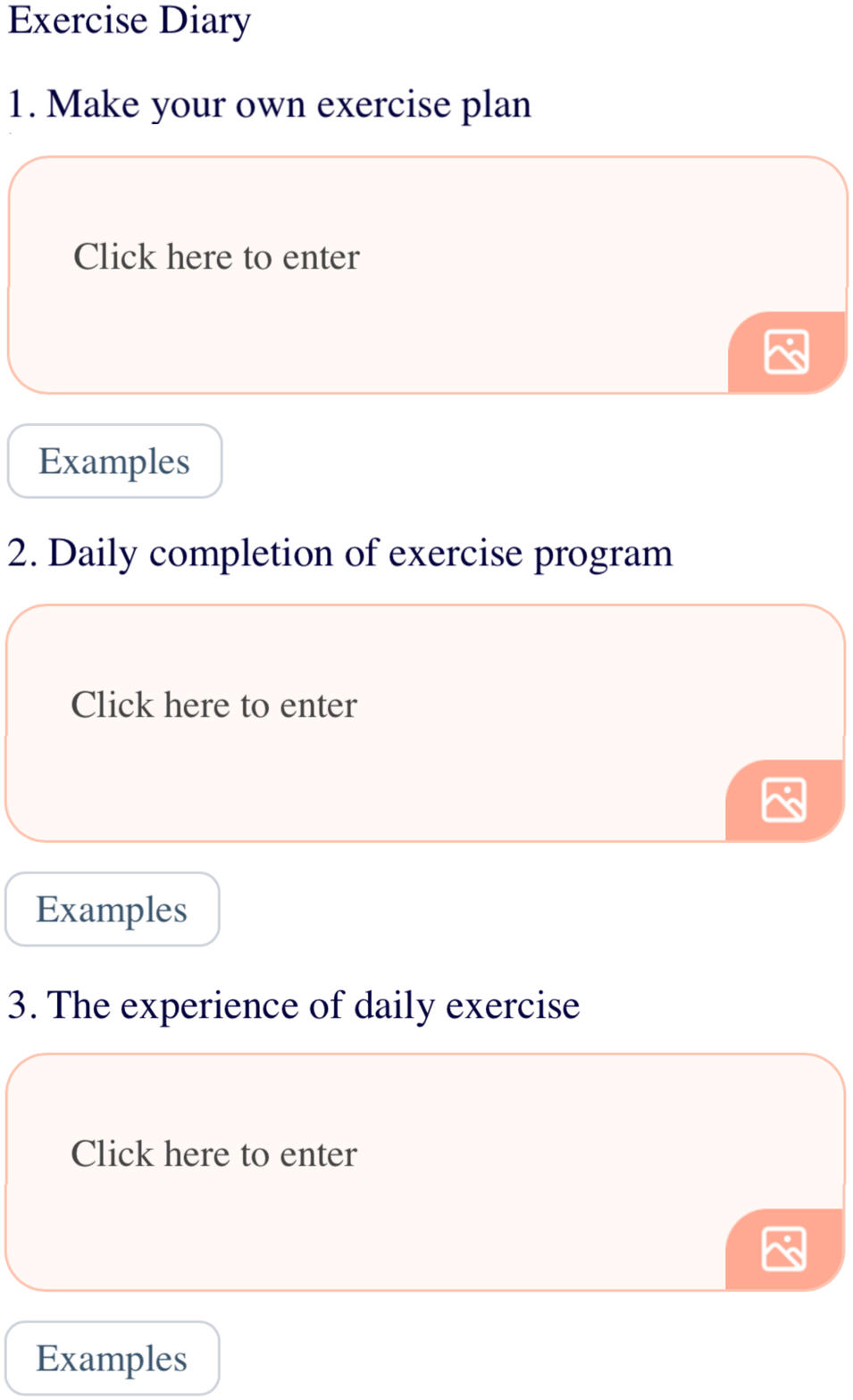}
    \caption{Guided Expressive Writing}
    \label{fig:cbt-diary}
\end{subfigure}
\caption{The user interface of Emohaa's CBT-based platform.}
\label{fig:cbt-first} 
\end{figure*}

\begin{figure*}[!ht]
\centering
\begin{subfigure}{.3\linewidth}
    \centering
    \includegraphics[width=\linewidth, height=4cm]{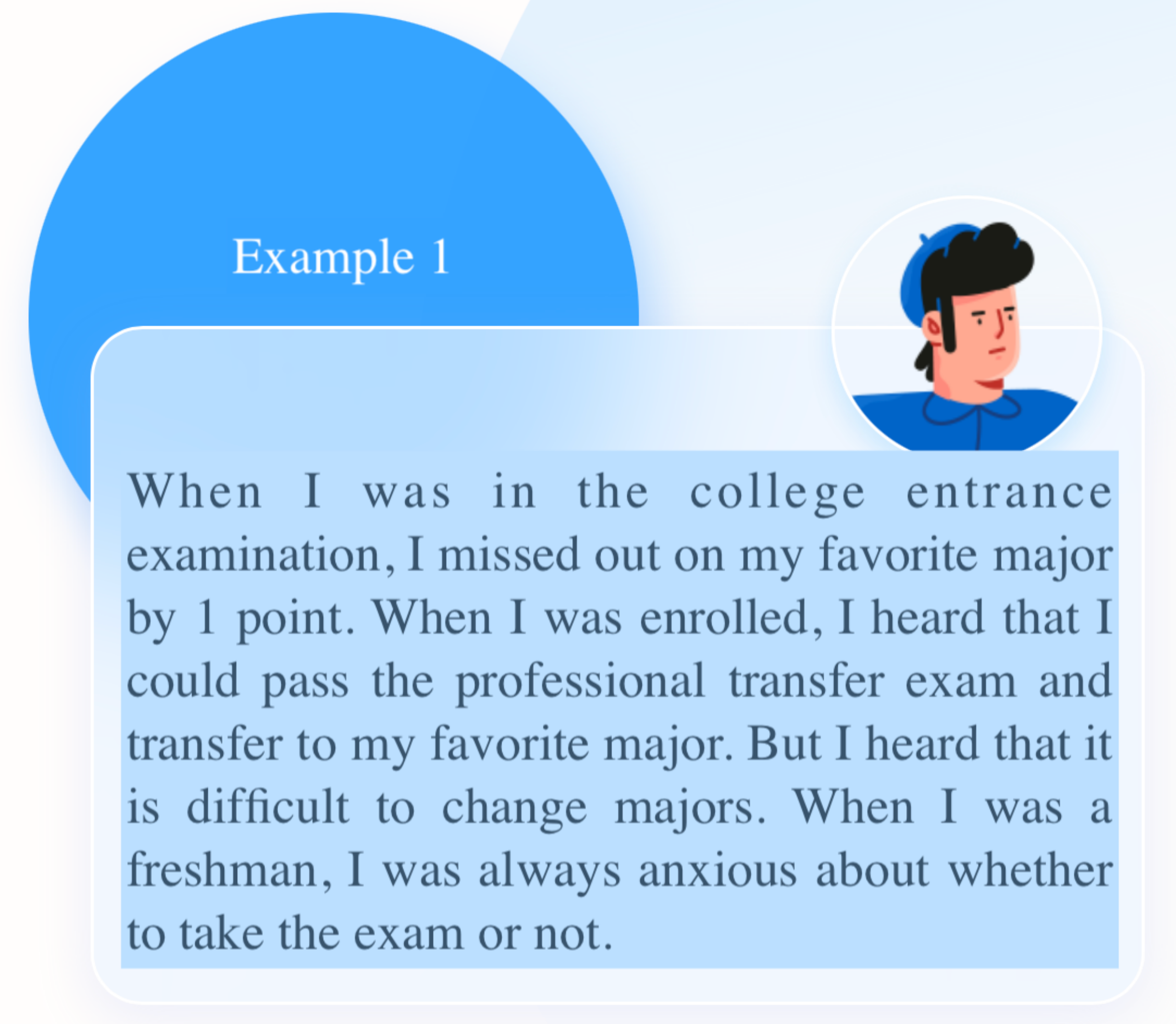}
    \caption{Step 1: Scenario Description}
\end{subfigure}
    \hfill
\begin{subfigure}{.3\linewidth}
    \centering
     \includegraphics[width=\linewidth, height=4cm]{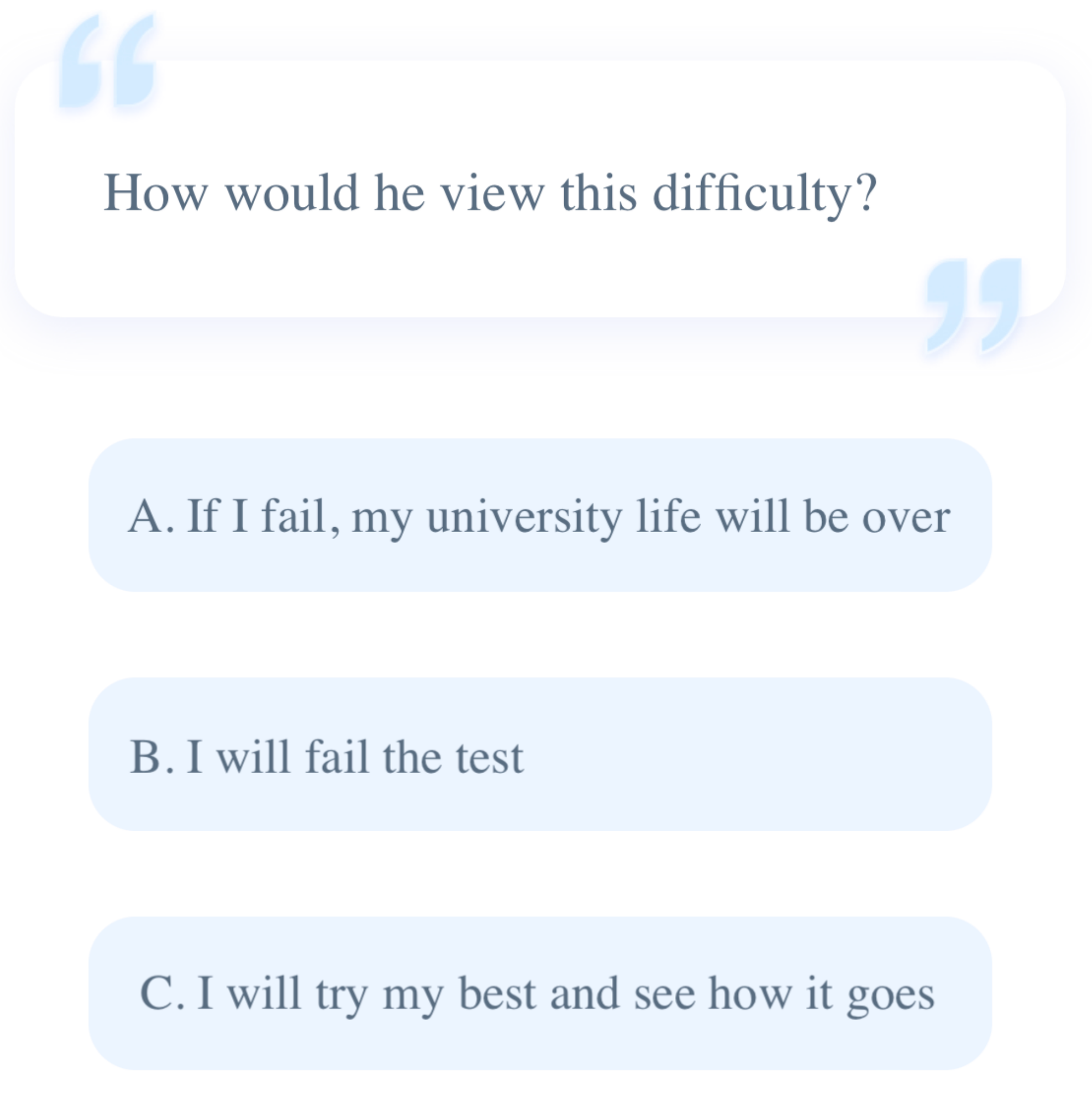}
     \caption{Step 2: Choosing a Reaction}
\end{subfigure}
    \hfill
\begin{subfigure}{.3\linewidth}
    \centering
     \includegraphics[width=\linewidth, height=4cm]{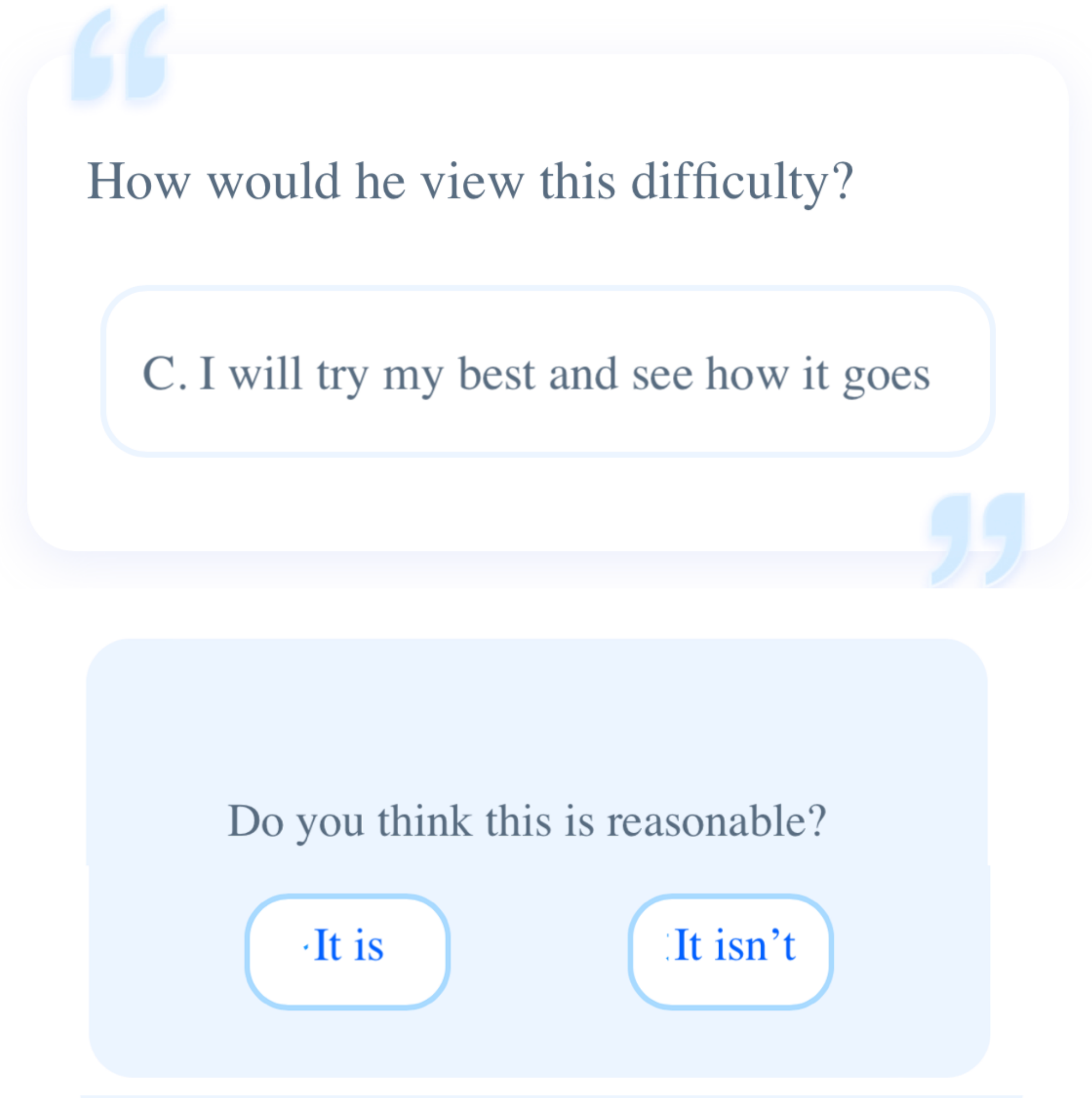}
     \caption{Step 3: Assessing Reasonability}
\end{subfigure}
\bigskip 
\vspace{0.1cm}
\begin{subfigure}{.3\linewidth}
    \centering
    \includegraphics[width=\linewidth, height=4cm]{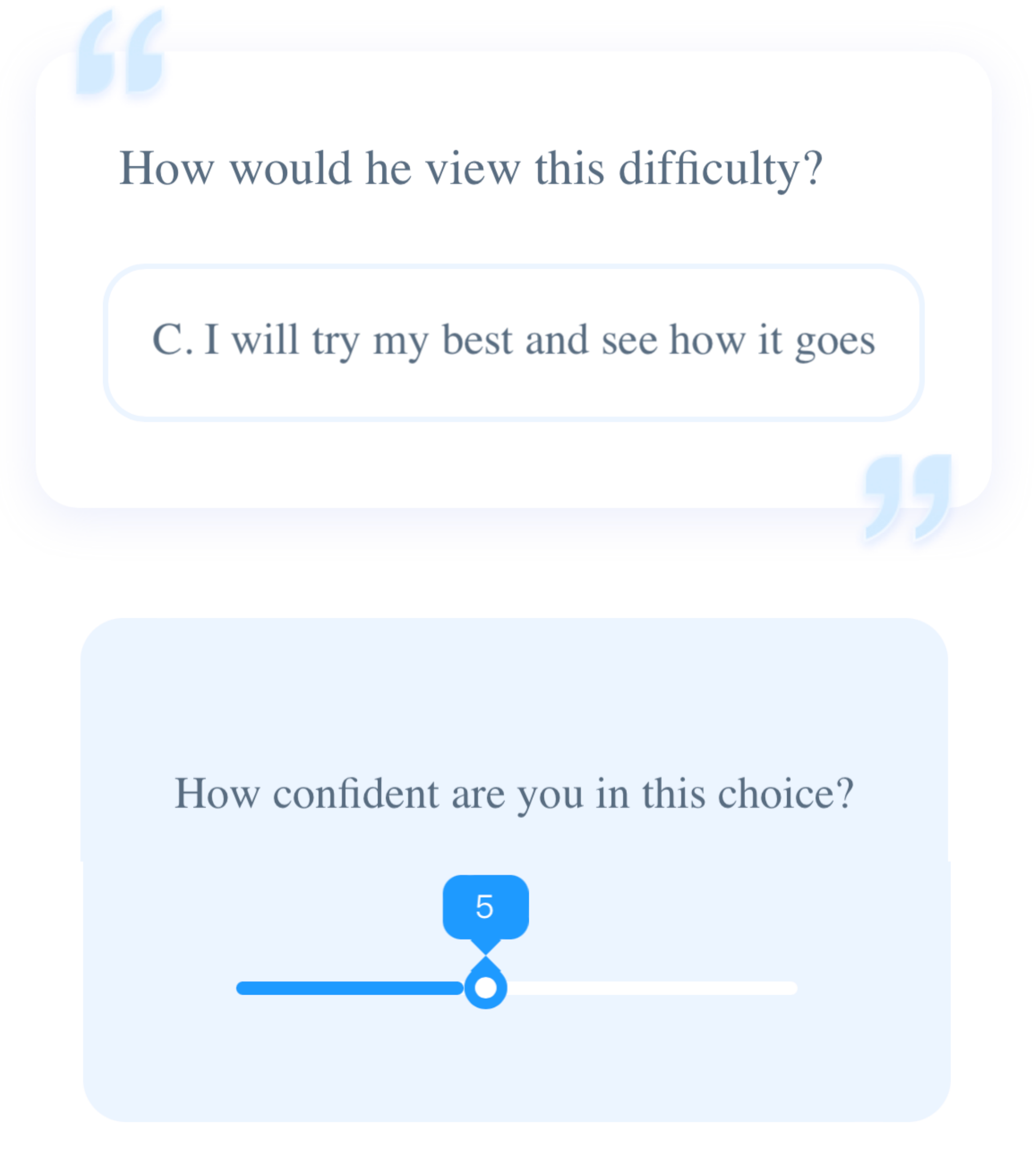}
    \caption{Step 4: Reporting Confidence}
\end{subfigure}
    \hfill
\begin{subfigure}{.3\linewidth}
    \centering
     \includegraphics[width=\linewidth, height=4cm]{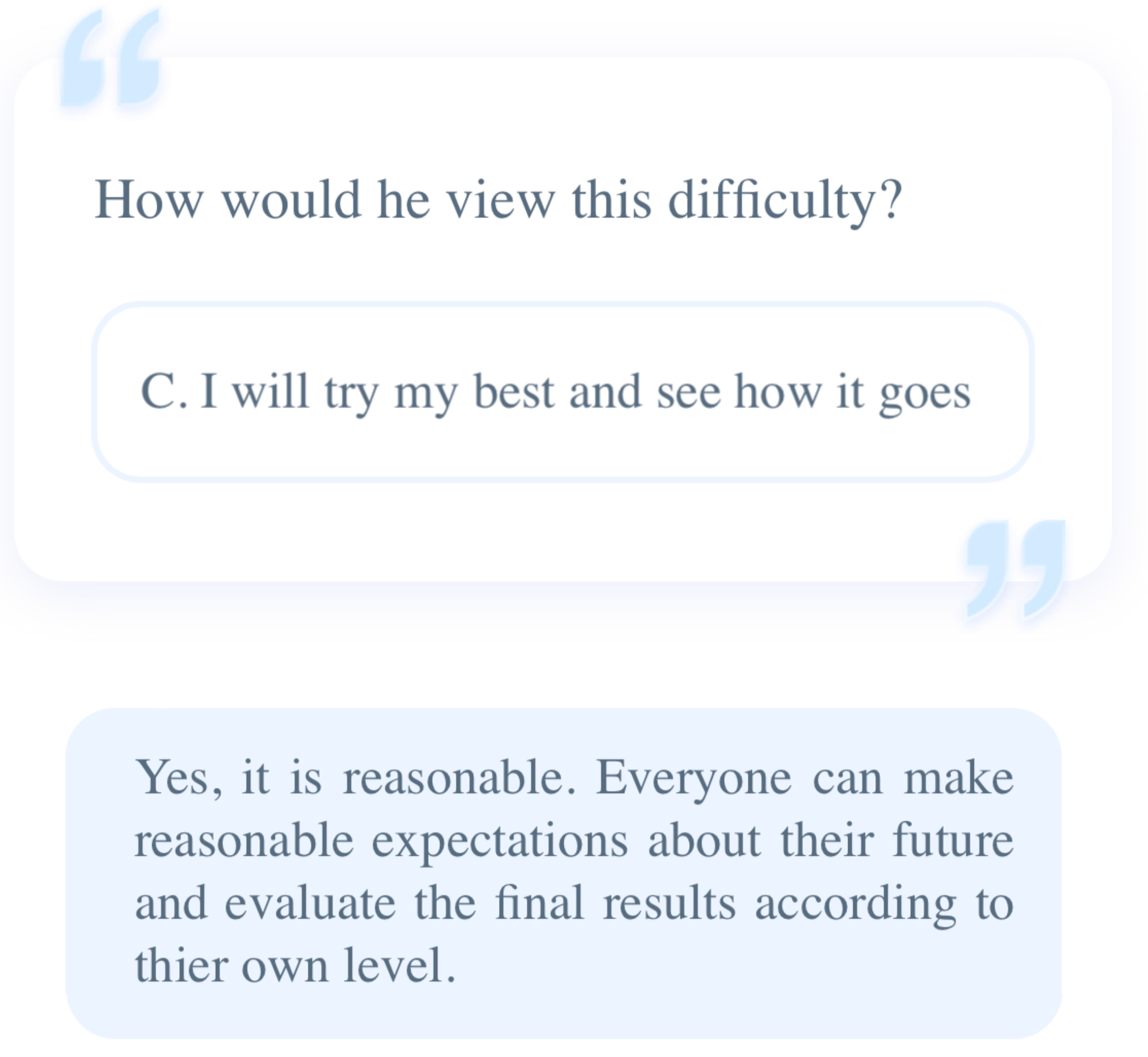}
     \caption{Step 5: Correct Perspective}
\end{subfigure}
    \hfill
\begin{subfigure}{.3\linewidth}
    \centering
     \includegraphics[width=\linewidth, height=4cm]{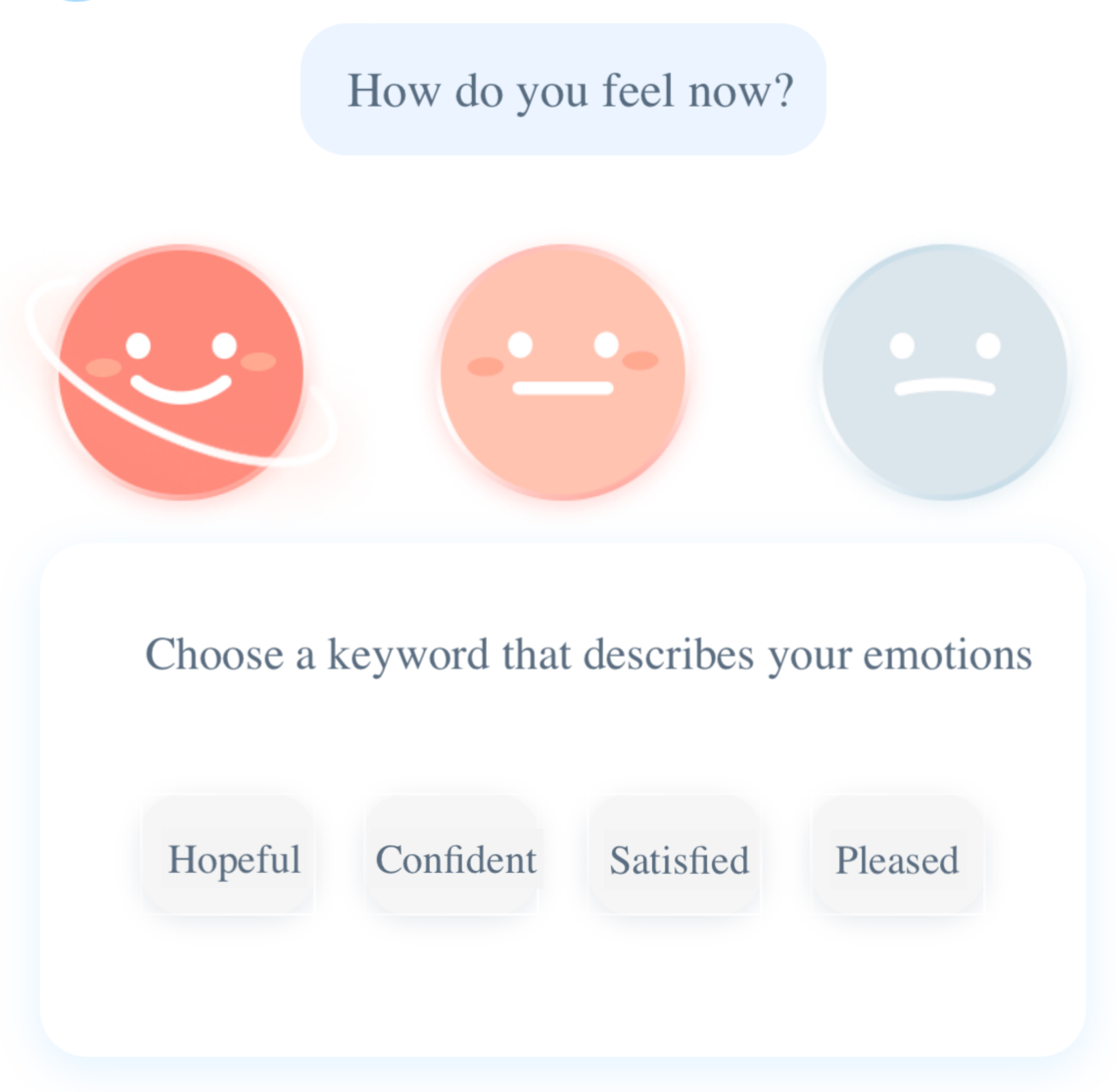}
     \caption{Step 6: Reporting Mood}
\end{subfigure}
\caption{An example of Automatic Thinking Exercises.}
\label{fig:cbt-at}
\end{figure*}

\begin{figure*}[!ht]
\centering
\begin{subfigure}{.45\linewidth}
    \centering
    \includegraphics[width=\linewidth, height=10cm]{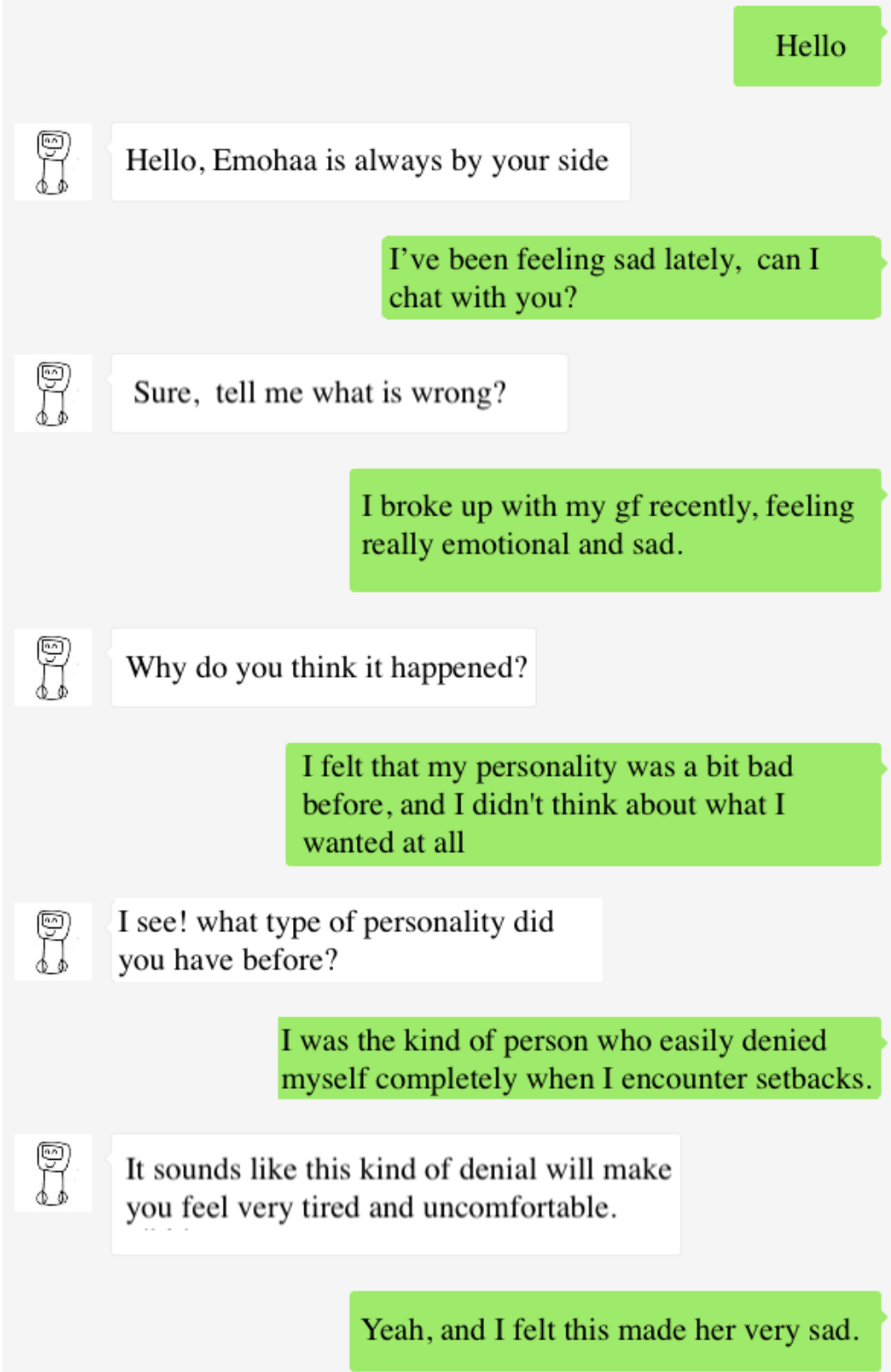}
\end{subfigure}
\hspace{0.2cm}
\begin{subfigure}{.45\linewidth}
    \centering
    \includegraphics[width=\linewidth, height=10cm]{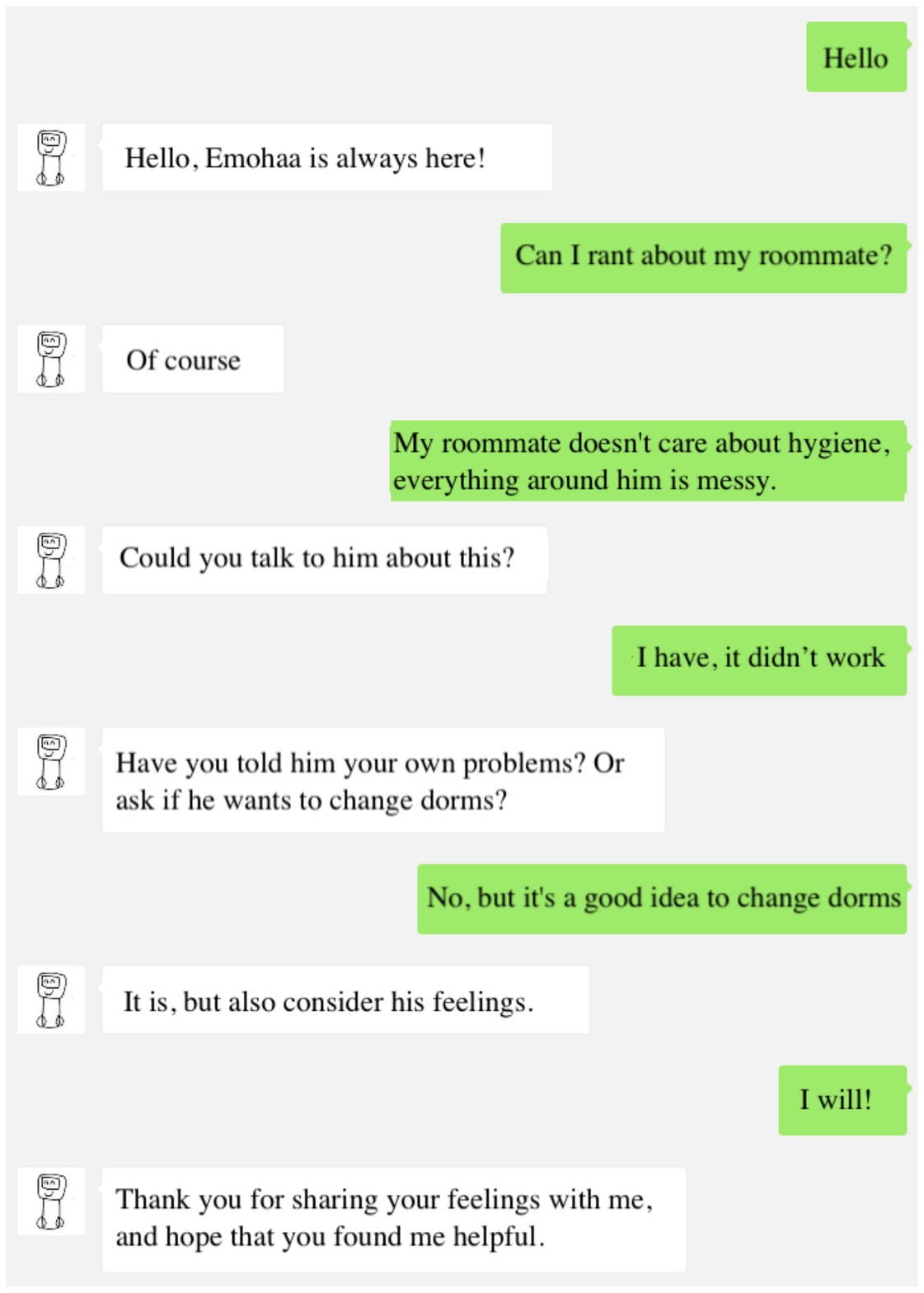}
\end{subfigure}
\caption{Example conversations with Emohaa's ES-based platform.}
\label{fig:es-chat}
\end{figure*}

Emohaa includes two platforms to provide cognitive and emotional support, respectively.
For cognitive support, we adopted CBT principles and designed the platform based on these practices. 
Participants are initially given a set of conversational choices on this platform and are accordingly introduced to CBT and how to use this platform (Figure \ref{fig:cbt_chat}). 
Emohaa's CBT-based platform provides two types of exercises: guided expressive writing and automatic thinking.
An example of a guided writing exercise is shown in Figure \ref{fig:cbt-diary}, where users are asked to fill out parts of their diary about a given topic in several steps.
Automatic thinking exercises (Figure \ref{fig:cbt-at}) present the user with a hypothetical scenario and require them to take the person's perspective in that situation. 
Accordingly, they are asked a question regarding the correct approach to take in that person's situation and report their confidence in their answer.
Lastly, they are shown the correct answer about how to approach and gain a new perspective in such scenarios.
To assess the efficacy of the exercises, we require users to report their mood after completing an exercise and describe their emotions using a set of pre-defined keywords.
Moreover, examples of emotional support on the ES-based platform, where users can freely discuss their emotional problems, are provided in Figure \ref{fig:es-chat}.
Notably, the user interfaces in the below figures were translated to English to facilitate general understanding for all readers.

\clearpage
\printcredits

\bibliographystyle{cas-model2-names}

\bibliography{cas-refs}

\end{document}